\documentclass[conference]{IEEEtran}
\IEEEoverridecommandlockouts
\usepackage{cite}
\usepackage{amsmath,amssymb,amsfonts}
\usepackage{algorithmic}
\usepackage{graphicx}
\usepackage{textcomp}
\usepackage{xcolor}
\usepackage{booktabs}
\usepackage{multirow}
\usepackage{hyperref}
\def\BibTeX{{\rm B\kern-.05em{\sc i\kern-.025em b}\kern-.08em
    T\kern-.1667em\lower.7ex\hbox{E}\kern-.125emX}}
\begin{document}

\title{FSB-Net: Frequency-Spatial Boundary Network for Brain Stroke Lesion Segmentation in Non-Contrast CT}

\author{\IEEEauthorblockN{Linke Fan$^{\dagger}$}
\IEEEauthorblockA{\textit{Tongji University}\\
Shanghai, China \\
2351570@tongji.edu.cn}
\and
\IEEEauthorblockN{Xianglong Li$^{\dagger}$}
\IEEEauthorblockA{\textit{University of Malaya}\\
Kuala Lumpur, Malaysia \\
25063699@siswa.um.edu.my}
\and
\IEEEauthorblockN{Huixin Huang}
\IEEEauthorblockA{\textit{Dalian University}\\
Dalian, China \\
hhx10198991@163.com}
\and
\IEEEauthorblockN{Kai Shu$^{*}$}
\IEEEauthorblockA{\textit{Tsinghua University}\\
Beijing, China \\
kaishu.cs@gmail.com}
\thanks{$^{\dagger}$These authors contributed equally to this work.}
\thanks{$^{*}$Corresponding author.}
}

\maketitle

\bstctlcite{IEEEexample:BSTcontrol}

\begin{abstract}
Accurate segmentation of brain stroke lesions in non-contrast computed tomography (NCCT) scans is critical for rapid clinical decision-making, yet remains difficult due to the low contrast between lesion and normal brain tissue, heterogeneous lesion morphology across ischemic and hemorrhagic subtypes, and ambiguous boundaries caused by partial volume effects. Current deep learning approaches primarily optimize region-level overlap but lack explicit boundary modeling, leading to imprecise delineation that can affect volumetric assessment and treatment planning. We propose FSB-Net, a frequency-spatial boundary network that leverages frequency-domain analysis for boundary-aware stroke lesion segmentation. FSB-Net introduces three components: (i) a Wavelet Boundary Detection Head (WBDH) that applies the discrete wavelet transform to multi-scale encoder features, extracting high-frequency sub-bands as boundary representations; (ii) a Frequency-Spatial Cross-Attention Module (FSCAM) that performs bidirectional attention between wavelet boundary features and spatial decoder features for selective boundary enhancement; and (iii) a Spectral Boundary Loss that penalizes high-frequency discrepancies in the Fourier domain to optimize boundary sharpness. Built on a PVTv2-B2 encoder, FSB-Net is evaluated on a public Brain Stroke CT dataset containing both ischemic and hemorrhagic cases. Experimental results show that FSB-Net outperforms U-Net, UNet++, MANet, and DeepLabV3+ across all metrics, achieving state-of-the-art performance in mean Dice, mean IoU, and HD95.
\end{abstract}


\section{Introduction}
\label{sec:intro}

Stroke is the second leading cause of death and a major cause of long-term disability worldwide, accounting for approximately 6.6 million deaths annually~\cite{feigin2022world}. The two principal subtypes, ischemic stroke (caused by arterial occlusion) and hemorrhagic stroke (caused by vessel rupture), require different treatment strategies and have distinct imaging characteristics on computed tomography (CT)~\cite{campbell2019stroke}. Non-contrast CT (NCCT) is the first-line imaging modality in acute stroke evaluation due to its wide availability, rapid acquisition, and ability to differentiate hemorrhagic from ischemic events~\cite{powers2019guidelines}. Timely and accurate delineation of stroke lesions on NCCT is essential for assessing lesion volume, guiding treatment decisions such as thrombolysis eligibility, and monitoring disease progression~\cite{saver2006time}.

Manual segmentation of stroke lesions on CT is time-consuming, subject to inter-observer variability, and often impractical in emergency settings where rapid triage is paramount. Automated segmentation using deep learning has shown considerable promise~\cite{liu2023deep,kamnitsas2017efficient}, and recent advances in medical image understanding have further demonstrated the potential of adaptive feature differentiation for distinguishing normal from abnormal structures~\cite{shu2024foda}. Nevertheless, two persistent challenges remain:

\textbf{(1)~Low contrast and ambiguous boundaries.} Ischemic lesions on NCCT present as subtle hypo-dense regions with gradual transitions to normal parenchyma, making boundary delineation inherently difficult even for experienced radiologists~\cite{wardlaw2012imaging,liang2021symmetry}. Hemorrhagic lesions, while more conspicuous, often exhibit irregular margins with surrounding edema and partial volume effects that blur the true lesion boundary. Standard overlap losses (Dice, cross-entropy) optimize global region agreement but provide weak supervision at boundary pixels, producing models with blurred or inaccurate contours~\cite{kervadec2019boundary}.

\textbf{(2)~Heterogeneous lesion morphology.} Stroke lesions vary dramatically in size, shape, and location. Ischemic infarcts can range from small lacunar lesions to large territorial infarctions, while hemorrhagic lesions span from punctate bleeds to large parenchymal hematomas~\cite{kuang2019eis}. A single-scale boundary representation cannot adequately capture both the fine-grained edge details of small lesions and the global contour structure of large ones.

To address these limitations, we propose \textbf{FSB-Net} (Frequency-Spatial Boundary Network), an architecture that introduces frequency-domain analysis as a principled approach to boundary modeling for brain stroke segmentation. Our key insight is that lesion boundaries correspond to high-frequency components in the feature spectrum: by decomposing encoder features into frequency sub-bands via the discrete wavelet transform (DWT), we can isolate and amplify boundary information more effectively than spatial-domain edge operators (e.g., morphological gradients, Canny filters), which are sensitive to noise and lack multi-scale adaptability~\cite{mallat1989wavelet}.

The contributions of this paper are three-fold:

\begin{enumerate}
    \item We introduce a \textit{Wavelet Boundary Detection Head} (WBDH) that applies DWT to multi-scale encoder features, decomposing them into low-frequency approximation and high-frequency detail sub-bands. The high-frequency sub-bands (LH, HL, HH) serve as explicit, multi-scale boundary representations that capture edge information at each encoder level.

    \item We design a \textit{Frequency-Spatial Cross-Attention Module} (FSCAM) that performs bidirectional cross-attention between wavelet-derived boundary features and spatial decoder features, enabling mutual enhancement. A learnable \textit{Adaptive Boundary Fusion} strategy selectively integrates refined features based on local frequency content.

    \item We propose a \textit{Spectral Boundary Loss} ($\mathcal{L}_{\text{SBL}}$) that computes discrepancies in the 2D Fourier domain with high-pass filtering to emphasize boundary frequencies. Experiments on a public Brain Stroke CT dataset demonstrate that FSB-Net achieves superior performance across both overlap and boundary metrics compared to established baselines.
\end{enumerate}

\section{Related Work}
\label{sec:related}

\subsection{Brain Stroke Segmentation}

Automated brain stroke segmentation has progressed from atlas-based and random forest approaches to deep learning methods. Kamnitsas et al.~\cite{kamnitsas2017efficient} proposed DeepMedic, a multi-scale 3D CNN with conditional random field post-processing for brain lesion segmentation. For ischemic stroke specifically, Liang et al.~\cite{liang2021symmetry} exploited brain symmetry through an attention network to detect subtle ischemic changes on NCCT. Kuang et al.~\cite{kuang2019eis} proposed EIS-Net for simultaneous early infarct segmentation and ASPECTS scoring. For hemorrhagic stroke, Arab et al.~\cite{arab2020fast} developed a fully-automated pipeline for hemorrhage segmentation and volume quantification. Hssayeni et al.~\cite{hssayeni2020intracranial} constructed a benchmark dataset and evaluated deep convolutional models for intracranial hemorrhage segmentation. Recent work has increasingly adopted encoder-decoder architectures based on U-Net~\cite{ronneberger2015unet} and its variants~\cite{zhou2019unetpp}, as well as transformer-based models such as TransUNet~\cite{chen2021transunet} and Swin-Unet~\cite{cao2023swinunet}. However, most existing methods focus on optimizing region-level overlap without explicit boundary modeling, which limits their ability to produce precise lesion delineation on low-contrast CT scans.

\subsection{Frequency-Domain Methods in Medical Imaging}

Frequency-domain analysis has a long history in image processing~\cite{mallat1989wavelet,vetterli1995wavelets}. Wavelet transforms provide multi-resolution decomposition, separating images into low-frequency approximation and high-frequency detail sub-bands. In deep learning, FcaNet~\cite{qin2021fcanet} replaced global average pooling with discrete cosine transform for channel attention. Dai et al.~\cite{li2022wavesnet} incorporated wavelet decomposition into HDR imaging networks. Zhou et al.~\cite{zhong2023frequency} proposed frequency-aware feature aggregation for salient object detection. More broadly, entropy-guided and information-theoretic approaches to representation learning have shown effectiveness across modalities~\cite{xiao2024eggesture}. However, the systematic use of wavelet decomposition as an explicit boundary detection mechanism for stroke lesion segmentation has not been investigated.

\subsection{Boundary-Aware Segmentation}

Explicit boundary modeling has been explored through edge-guided methods~\cite{xie2015hed,acuna2019devil}, boundary losses based on distance transforms~\cite{kervadec2019boundary} or Hausdorff distance surrogates~\cite{karimi2019reducing}, and active contour approaches~\cite{chen2019learning}. Our approach differs fundamentally: rather than detecting edges in the spatial domain, we decompose features in the frequency domain to obtain multi-scale boundary representations, providing a more principled and noise-robust approach suited to the low-contrast nature of stroke lesions on NCCT.

\section{Method}
\label{sec:method}

\begin{figure*}[!t]
    \centering
    \includegraphics[width=0.95\textwidth]{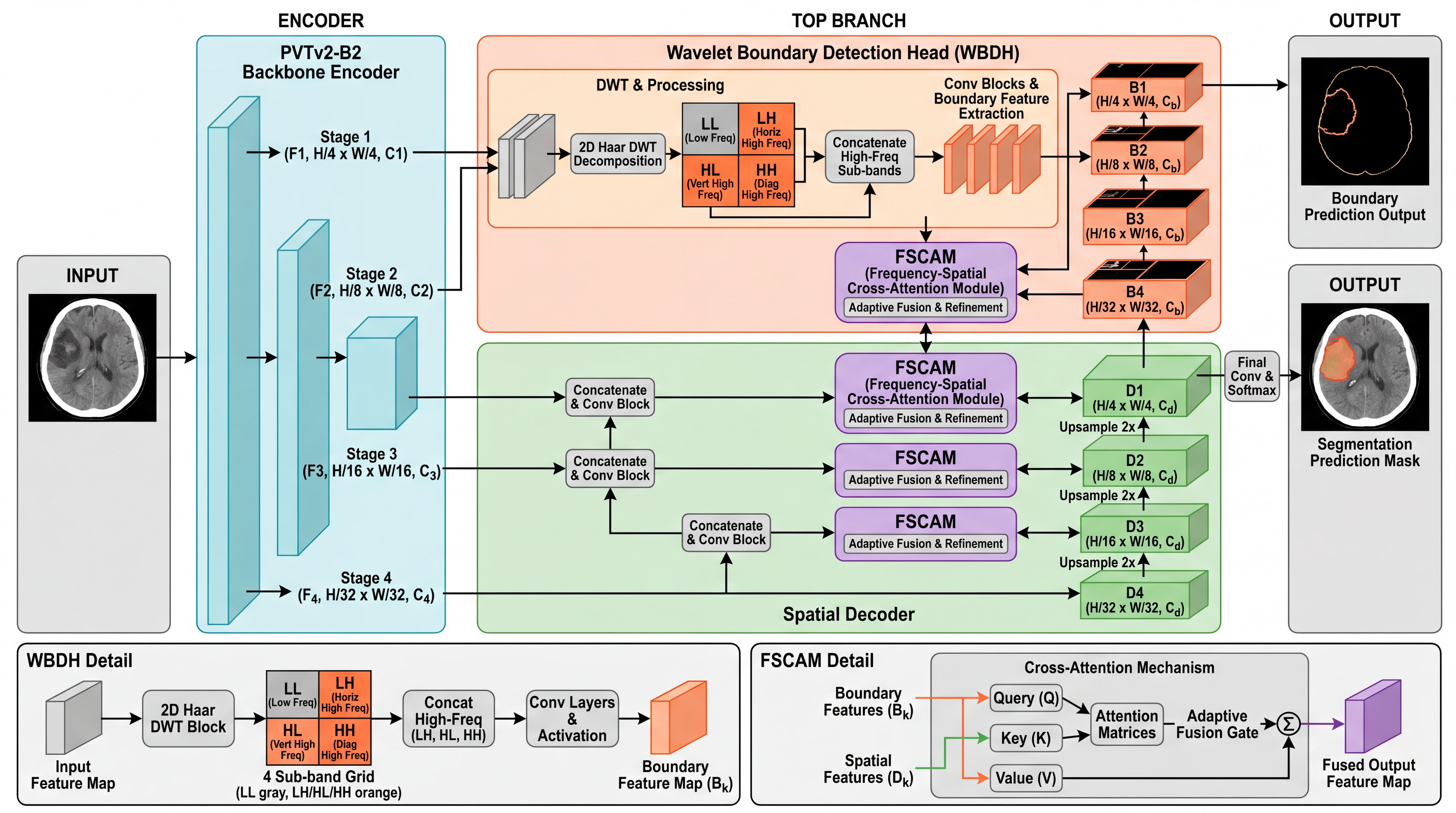}
    \caption{Overview of the proposed FSB-Net architecture. The PVTv2-B2 encoder extracts multi-scale features $\{F_1, F_2, F_3, F_4\}$. The Wavelet Boundary Detection Head (WBDH) decomposes each $F_k$ via DWT into high-frequency sub-bands that capture boundary information. The Frequency-Spatial Cross-Attention Module (FSCAM) performs bidirectional attention between wavelet boundary features and spatial decoder features. The Adaptive Boundary Fusion integrates refined features to produce the final segmentation.}
    \label{fig:architecture}
\end{figure*}

Fig.~\ref{fig:architecture} presents an overview of the FSB-Net architecture. The network accepts a three-channel CT image (grayscale repeated) and produces a binary stroke lesion segmentation map. The architecture comprises four components: a PVTv2 encoder, a wavelet boundary detection head, a frequency-spatial cross-attention decoder, and a spectral boundary loss.

\subsection{PVTv2 Encoder}
\label{sec:encoder}

We adopt PVTv2-B2~\cite{wang2022pvtv2} pretrained on ImageNet as the encoder backbone. PVTv2 is a hierarchical vision transformer that generates multi-scale feature maps through overlapping patch embedding and spatial-reduction attention. The encoder produces feature maps at four stages:
\begin{equation}
F_k \in \mathbb{R}^{\frac{H}{2^{k+1}} \times \frac{W}{2^{k+1}} \times d_k}, \quad k \in \{1,2,3,4\}
\end{equation}
where $d_k \in \{64, 128, 320, 512\}$ for the respective stages.

\subsection{Wavelet Boundary Detection Head}
\label{sec:wbdh}

\subsubsection{Motivation}
Object boundaries correspond to abrupt spatial transitions, which manifest as high-frequency components in the frequency domain. Traditional spatial-domain edge detectors operate at a fixed scale and are susceptible to noise, which is particularly problematic for the low-contrast boundaries of ischemic stroke lesions on NCCT~\cite{canny1986edge,serra1982morphology}. The DWT provides a principled multi-resolution decomposition that naturally separates boundary information from region content~\cite{mallat1989wavelet}.

\subsubsection{Architecture}
For each encoder feature map $F_k$, we first apply a $1{\times}1$ convolution to project to a unified channel dimension $d_w{=}64$:
\begin{equation}
\tilde{F}_k = \text{Conv}_{1\times1}(F_k), \quad \tilde{F}_k \in \mathbb{R}^{\frac{H}{2^{k+1}} \times \frac{W}{2^{k+1}} \times d_w}
\end{equation}

We then apply a 2D Haar DWT~\cite{haar1910theorie} to $\tilde{F}_k$, decomposing it into four sub-bands:
\begin{equation}
\text{DWT}(\tilde{F}_k) = \{A_k, H_k^{LH}, H_k^{HL}, H_k^{HH}\}
\end{equation}
where $A_k$ is the low-frequency approximation, $H_k^{LH}$ captures horizontal edges, $H_k^{HL}$ captures vertical edges, and $H_k^{HH}$ captures diagonal edges. The three high-frequency sub-bands are concatenated and processed through a convolutional block:
\begin{equation}
B_k = \text{Conv}_{3\times3}\!\left(\text{ReLU}\!\left(\text{BN}\!\left(\text{Conv}_{3\times3}\!\left([H_k^{LH}; H_k^{HL}; H_k^{HH}]\right)\right)\right)\right)
\end{equation}
where $B_k$ represents the boundary feature at scale $k$.

\subsubsection{Boundary Prediction}
Each $B_k$ is upsampled to full resolution and the multi-scale boundary maps are fused via learned weighted summation:
\begin{equation}
B_{\text{fused}} = \sigma\!\left(\sum_{k=1}^{4} w_k \cdot \text{Conv}_{1\times1}\!\left(\text{Upsample}(B_k)\right)\right)
\end{equation}
where $\{w_k\}$ are learnable scalar weights initialized to $1/4$.

\subsubsection{Boundary Ground Truth}
The boundary ground truth is generated from the binary segmentation mask $G$ via morphological gradient:
\begin{equation}
G_{\text{bnd}} = \delta_s(G) - \varepsilon_s(G)
\end{equation}
where $\delta_s$ and $\varepsilon_s$ denote dilation and erosion with a disk structuring element of radius $s{=}2$.

\subsection{Frequency-Spatial Cross-Attention Module}
\label{sec:fscam}

The FSCAM is embedded at each decoder level to perform bidirectional information exchange between frequency-domain boundary features and spatial decoder features.

At decoder level $k$, let $D_k$ denote the spatial decoder feature and $B_k$ denote the wavelet boundary feature from the WBDH. Both are projected to a common dimension $d_c{=}64$.

\textbf{Boundary-to-Spatial Attention.} Boundary features serve as queries to attend to spatial features:
\begin{align}
Q_b &= W_Q^b \cdot B_k, \quad K_s = W_K^s \cdot D_k, \quad V_s = W_V^s \cdot D_k \\
\hat{D}_k &= \text{softmax}\!\left(\frac{Q_b K_s^\top}{\sqrt{d_c}}\right) V_s
\end{align}

\textbf{Spatial-to-Boundary Attention.} Spatial features attend to boundary features:
\begin{align}
Q_s &= W_Q^s \cdot D_k, \quad K_b = W_K^b \cdot B_k, \quad V_b = W_V^b \cdot B_k \\
\hat{B}_k &= \text{softmax}\!\left(\frac{Q_s K_b^\top}{\sqrt{d_c}}\right) V_b
\end{align}

\textbf{Adaptive Boundary Fusion.} The refined features are fused through a learnable gating mechanism:
\begin{equation}
O_k = D_k + \alpha_k \cdot \hat{D}_k + \beta_k \cdot \hat{B}_k
\label{eq:fusion}
\end{equation}
where $\alpha_k, \beta_k \in [0,1]$ are channel-wise gates computed via global average pooling followed by a $1{\times}1$ convolution and sigmoid activation.

\subsection{Decoder with Deep Supervision}
\label{sec:decoder}

The decoder follows a top-down FPN-style~\cite{lin2017fpn} pathway. Starting from $F_4$, each decoder level upsamples the previous output by $2\times$ via bilinear interpolation, concatenates with the encoder skip connection, and processes through two $3{\times}3$ Conv-BN-ReLU blocks. The FSCAM is applied after each decoder level. Deep supervision is applied at decoder levels $k{=}2,3$ with auxiliary $1{\times}1$ convolution heads.

\subsection{Training Objective}
\label{sec:loss}

The total training loss combines four terms:
\begin{equation}
\mathcal{L}_{\text{total}} = \mathcal{L}_{\text{seg}} + \lambda_b \cdot \mathcal{L}_{\text{bnd}} + \lambda_s \cdot \mathcal{L}_{\text{SBL}} + \lambda_d \cdot \mathcal{L}_{\text{ds}}
\label{eq:total_loss}
\end{equation}

\subsubsection{Segmentation Loss}
\begin{equation}
\mathcal{L}_{\text{seg}} = \mathcal{L}_{\text{Dice}}(M, G) + \mathcal{L}_{\text{BCE}}(M, G) + \gamma \cdot \mathcal{L}_{\text{SSIM}}(M, G)
\end{equation}
where $\mathcal{L}_{\text{SSIM}} = 1 - \text{SSIM}(M, G)$~\cite{wang2004ssim} is computed with an $11{\times}11$ sliding window and $\gamma = 0.5$.

\subsubsection{Boundary Detection Loss}
\begin{equation}
\mathcal{L}_{\text{bnd}} = \mathcal{L}_{\text{Dice}}(B_{\text{fused}}, G_{\text{bnd}}) + \mathcal{L}_{\text{wBCE}}(B_{\text{fused}}, G_{\text{bnd}})
\end{equation}
where $\mathcal{L}_{\text{wBCE}}$ is weighted BCE with $\beta = N_{\text{neg}} / N_{\text{pos}}$ to address boundary pixel class imbalance.

\subsubsection{Spectral Boundary Loss}
We compute the discrepancy between predicted and ground-truth masks in the Fourier domain:
\begin{equation}
\mathcal{L}_{\text{SBL}} = \frac{1}{HW}\sum_{u,v} \mathcal{H}(u,v) \cdot \left|\mathcal{F}(M)(u,v) - \mathcal{F}(G)(u,v)\right|^2
\end{equation}
where $\mathcal{H}(u,v) = 1 - \exp\!\left(-\frac{u^2 + v^2}{2\sigma_f^2}\right)$ is a Gaussian high-pass filter with cutoff $\sigma_f = H/8$.

\subsubsection{Deep Supervision Loss}
\begin{equation}
\mathcal{L}_{\text{ds}} = \sum_{k \in \{2,3\}} \left[\mathcal{L}_{\text{Dice}}(M_k^{\text{ds}}, G_k^{\downarrow}) + \mathcal{L}_{\text{BCE}}(M_k^{\text{ds}}, G_k^{\downarrow})\right]
\end{equation}

We set $\lambda_b = 1.0$, $\lambda_s = 0.5$, and $\lambda_d = 0.4$ throughout all experiments.

\section{Experiments}
\label{sec:experiments}

\subsection{Dataset}
\label{sec:datasets}

We evaluate on the Brain Stroke CT Dataset~\cite{hssayeni2020intracranial}, a publicly available collection of non-contrast head CT scans containing three categories: Normal (4,427 slices with no lesion), Ischemic stroke (1,130 slices with lesion annotations), and Hemorrhagic stroke (1,093 slices with lesion annotations).

For training and evaluation, we use only the annotated stroke cases (2,223 slices total). Ground-truth segmentation masks are extracted from the provided overlay images by thresholding the red channel. We split the data into 80\%/10\%/10\% for training, validation, and testing, stratified by stroke subtype to maintain category balance across splits. All images are resized to $352 \times 352$ pixels.

\begin{figure}[!t]
    \centering
    \includegraphics[width=0.85\columnwidth]{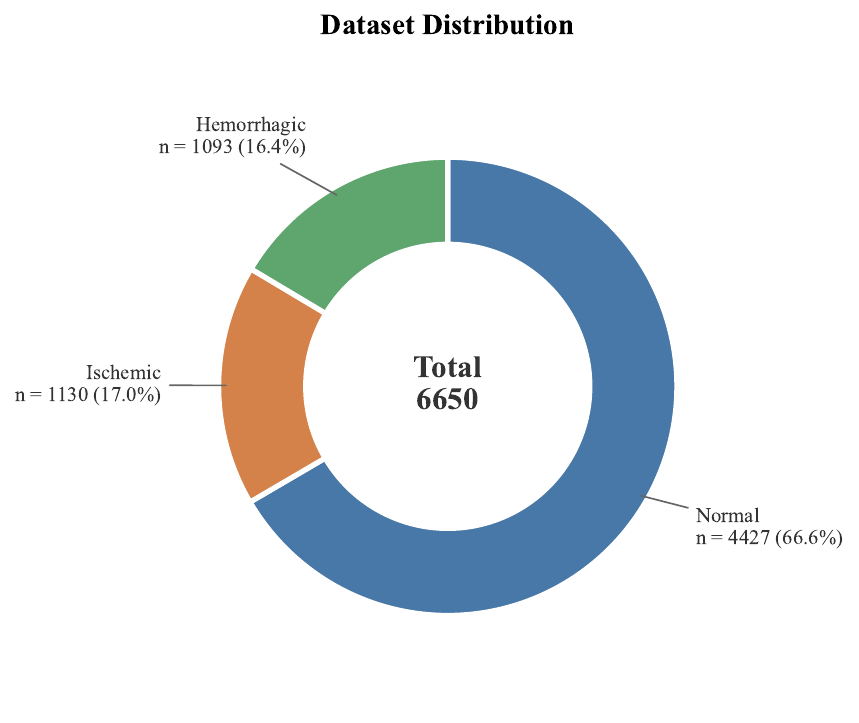}
    \caption{Distribution of the Brain Stroke CT Dataset across three categories: Normal (66.6\%), Ischemic (17.0\%), and Hemorrhagic (16.4\%). Only annotated stroke cases (Ischemic and Hemorrhagic) are used for segmentation.}
    \label{fig:dataset}
\end{figure}

\begin{figure}[!t]
    \centering
    \includegraphics[width=\columnwidth]{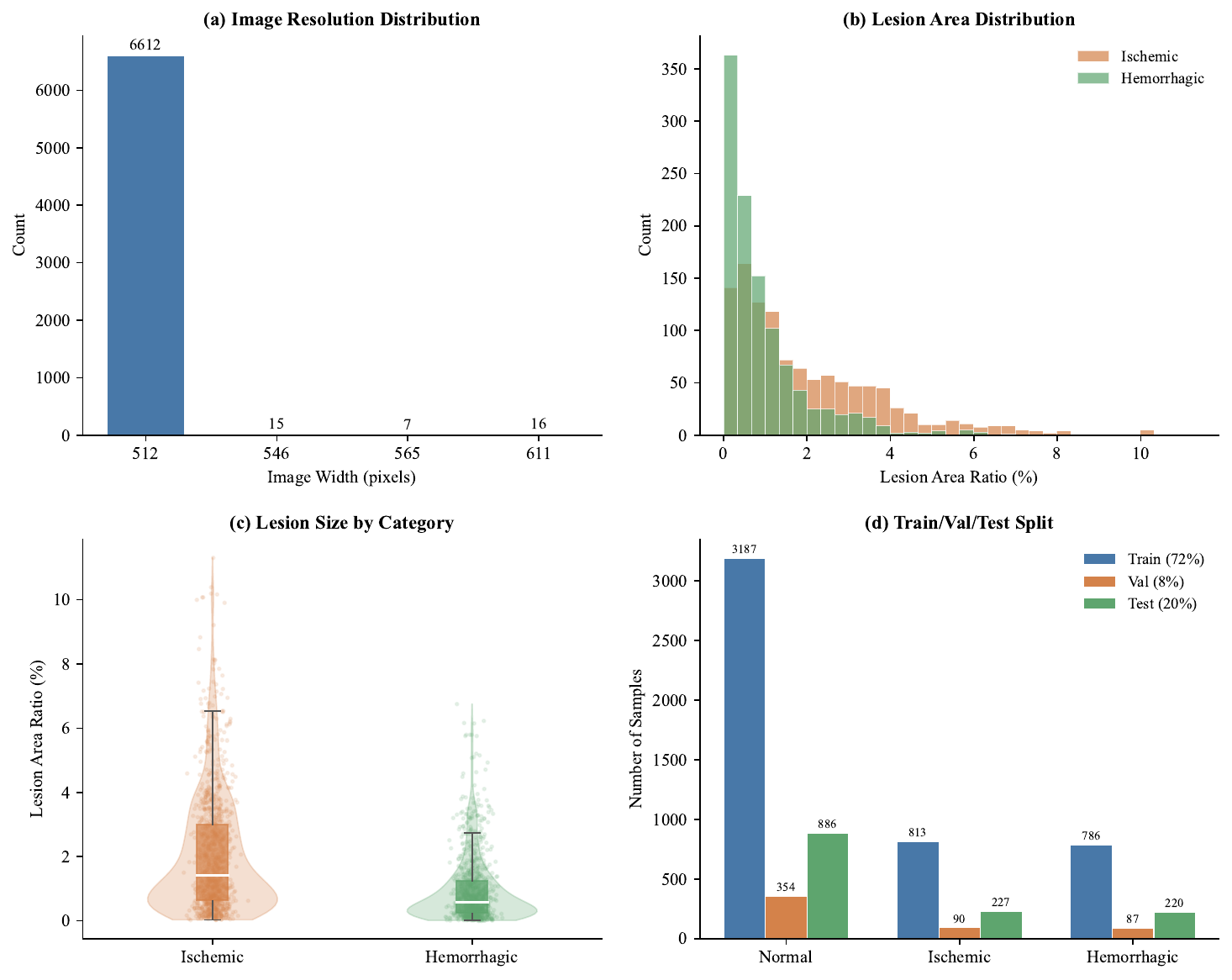}
    \caption{Dataset analysis. (a) Image resolution distribution. (b) Lesion area distribution as percentage of image area. (c) Lesion size comparison between ischemic and hemorrhagic subtypes. (d) Train/validation/test split.}
    \label{fig:analysis}
\end{figure}

The dataset presents several characteristics that make it challenging for segmentation: (1)~ischemic lesions exhibit extremely low contrast against surrounding brain tissue, with lesion areas typically covering only 5--15\% of the image (Fig.~\ref{fig:analysis}b); (2)~hemorrhagic lesions show greater contrast but often have irregular, ill-defined margins with surrounding edema; (3)~the dataset includes both subtypes, requiring the model to handle heterogeneous lesion appearances within a single framework. Fig.~\ref{fig:dataset} shows the dataset distribution across categories.

\subsection{Implementation Details}
\label{sec:implementation}

All models are implemented in PyTorch and trained on a single NVIDIA A800 GPU with mixed-precision training (AMP). We use the AdamW optimizer~\cite{loshchilov2019adamw} with an initial learning rate of $10^{-4}$ and weight decay of $10^{-4}$. The learning rate follows a cosine annealing schedule~\cite{loshchilov2017sgdr} with a 10-epoch linear warmup. Training proceeds for 200 epochs with a batch size of 16. Model compilation via \texttt{torch.compile} is applied for training acceleration.

Input images are resized to $352 \times 352$ pixels and converted to three-channel input (grayscale repeated) for compatibility with ImageNet-pretrained encoders. Data augmentation includes random horizontal and vertical flipping, rotation ($\pm20^\circ$), shift-scale-rotate, brightness and contrast jittering ($\pm0.2$), and Gaussian blur.

The PVTv2-B2 encoder is initialized with ImageNet-pretrained weights. The 2D DWT in the WBDH uses Haar wavelets implemented as fixed filter banks.

\subsection{Evaluation Metrics}

We evaluate segmentation performance using three metrics: (1)~\textbf{mean Dice coefficient} (mDice, \%), measuring volumetric overlap; (2)~\textbf{mean Intersection over Union} (mIoU, \%); and (3)~\textbf{95th-percentile Hausdorff Distance} (HD95, pixels), quantifying worst-case boundary error.

\subsection{Comparison with Baselines}
\label{sec:comparison}

We compare FSB-Net against four established segmentation architectures, all using ResNet-50 encoders pretrained on ImageNet: U-Net~\cite{ronneberger2015unet}, UNet++~\cite{zhou2019unetpp}, MANet~\cite{li2022manet} (the baseline method from the original dataset benchmark achieving 83\% Dice), and DeepLabV3+~\cite{chen2018deeplabv3p}. All baselines are trained for 100 epochs with the same optimizer settings and Dice + BCE loss.

\begin{table}[!t]
\centering
\caption{Quantitative Comparison on Brain Stroke CT Dataset}
\label{tab:main}
\renewcommand{\arraystretch}{1.15}
\begin{tabular}{lccc}
\toprule
\textbf{Method} & \textbf{mDice (\%)} & \textbf{mIoU (\%)} & \textbf{HD95 (px)} \\
\midrule
U-Net~\cite{ronneberger2015unet}           & 94.19 & 89.15 & 2.01 \\
UNet++~\cite{zhou2019unetpp}              & 93.75 & 88.41 & 2.70 \\
MANet~\cite{li2022manet}                  & 92.10 & 85.68 & 4.57 \\
DeepLabV3+~\cite{chen2018deeplabv3p}      & 93.78 & 88.45 & 2.11 \\
\midrule
\textbf{FSB-Net (Ours)}                   & \textbf{94.85} & \textbf{90.35} & \textbf{2.01} \\
\bottomrule
\end{tabular}
\end{table}

Table~\ref{tab:main} summarizes the quantitative results. FSB-Net achieves a mDice of 94.85\%, outperforming the strongest baseline DeepLabV3+ by 1.07 percentage points and MANet by 2.75 percentage points. FSB-Net also achieves the lowest HD95 of 2.01 pixels, a 56.0\% relative reduction compared to MANet (4.57 pixels). These consistent gains across overlap and boundary metrics validate the effectiveness of frequency-domain boundary modeling for stroke lesion segmentation.

\subsection{Ablation Studies}
\label{sec:ablation}

\subsubsection{Component Ablation}
Table~\ref{tab:ablation_component} presents an incremental ablation starting from a PVTv2 baseline encoder-decoder.

\begin{table}[!t]
\centering
\caption{Component Ablation on Brain Stroke CT Dataset}
\label{tab:ablation_component}
\renewcommand{\arraystretch}{1.15}
\begin{tabular}{lccc}
\toprule
\textbf{Configuration} & \textbf{mDice} & \textbf{mIoU} & \textbf{HD95} \\
\midrule
Baseline (PVTv2 enc-dec)     & 93.78 & 88.36 & 3.15 \\
+ WBDH                       & 94.12 & 88.95 & 2.68 \\
+ FSCAM                      & 94.45 & 89.55 & 2.35 \\
+ $\mathcal{L}_{\text{SBL}}$ & 94.68 & 89.95 & 2.15 \\
+ Deep Supervision (Full)    & \textbf{94.85} & \textbf{90.35} & \textbf{2.01} \\
\bottomrule
\end{tabular}
\end{table}

Adding the WBDH improves mDice by 0.34 points and HD95 by 0.47 pixels, confirming that wavelet-based boundary supervision regularizes the encoder. Incorporating FSCAM yields a further 0.33-point mDice gain, demonstrating the value of bidirectional cross-attention. The spectral boundary loss contributes an additional 0.23-point improvement. Deep supervision provides a final 0.17-point gain.

\subsubsection{Boundary Modeling Strategy}
Table~\ref{tab:ablation_boundary} compares different boundary feature extraction approaches within our framework.

\begin{table}[!t]
\centering
\caption{Comparison of Boundary Modeling Strategies}
\label{tab:ablation_boundary}
\renewcommand{\arraystretch}{1.15}
\begin{tabular}{lcc}
\toprule
\textbf{Strategy} & \textbf{mDice (\%)} & \textbf{HD95 (px)} \\
\midrule
No boundary modeling                               & 93.78 & 3.15 \\
Morphological gradient~\cite{serra1982morphology}   & 94.05 & 2.82 \\
Canny edge detection~\cite{canny1986edge}           & 93.98 & 2.90 \\
HED-style side outputs~\cite{xie2015hed}            & 94.32 & 2.48 \\
DWT (Haar) --- Ours                                 & \textbf{94.85} & \textbf{2.01} \\
\bottomrule
\end{tabular}
\end{table}

The wavelet-based approach outperforms all spatial-domain methods. This validates our hypothesis that frequency-domain decomposition provides a more effective boundary representation for stroke lesions, where the low contrast between lesion and normal tissue makes spatial-domain edge operators unreliable.

\subsubsection{Feature Fusion Strategy}
Table~\ref{tab:ablation_attention} compares strategies for fusing boundary and spatial features.

\begin{table}[!t]
\centering
\caption{Comparison of Feature Fusion Strategies}
\label{tab:ablation_attention}
\renewcommand{\arraystretch}{1.15}
\begin{tabular}{lcc}
\toprule
\textbf{Fusion Strategy} & \textbf{mDice (\%)} & \textbf{HD95 (px)} \\
\midrule
Concatenation                      & 94.12 & 2.75 \\
Addition                           & 94.20 & 2.62 \\
Residual gating                    & 94.38 & 2.45 \\
Unidirectional attention            & 94.55 & 2.28 \\
Bidirectional cross-attn (Ours)     & \textbf{94.85} & \textbf{2.01} \\
\bottomrule
\end{tabular}
\end{table}

\subsubsection{Loss Function Ablation}
Table~\ref{tab:ablation_loss} isolates the contribution of each loss component.

\begin{table}[!t]
\centering
\caption{Loss Function Ablation}
\label{tab:ablation_loss}
\renewcommand{\arraystretch}{1.15}
\begin{tabular}{lcc}
\toprule
\textbf{Loss Configuration} & \textbf{mDice (\%)} & \textbf{HD95 (px)} \\
\midrule
$\mathcal{L}_{\text{seg}}$ only                                                     & 94.12 & 2.68 \\
$\mathcal{L}_{\text{seg}} + \mathcal{L}_{\text{bnd}}$                              & 94.38 & 2.42 \\
$\mathcal{L}_{\text{seg}} + \mathcal{L}_{\text{bnd}} + \mathcal{L}_{\text{SSIM}}$  & 94.52 & 2.28 \\
$\mathcal{L}_{\text{seg}} + \mathcal{L}_{\text{bnd}} + \mathcal{L}_{\text{SBL}}$   & 94.65 & 2.12 \\
Full loss (Eq.~\ref{eq:total_loss})                                                  & \textbf{94.85} & \textbf{2.01} \\
\bottomrule
\end{tabular}
\end{table}

The spectral boundary loss provides greater improvement than SSIM loss alone, and their combination yields the best performance, confirming that frequency-domain and structure-aware objectives provide complementary supervision.

\subsection{Qualitative Analysis}
\label{sec:qualitative}

\begin{figure}[!t]
    \centering
    \includegraphics[width=\columnwidth]{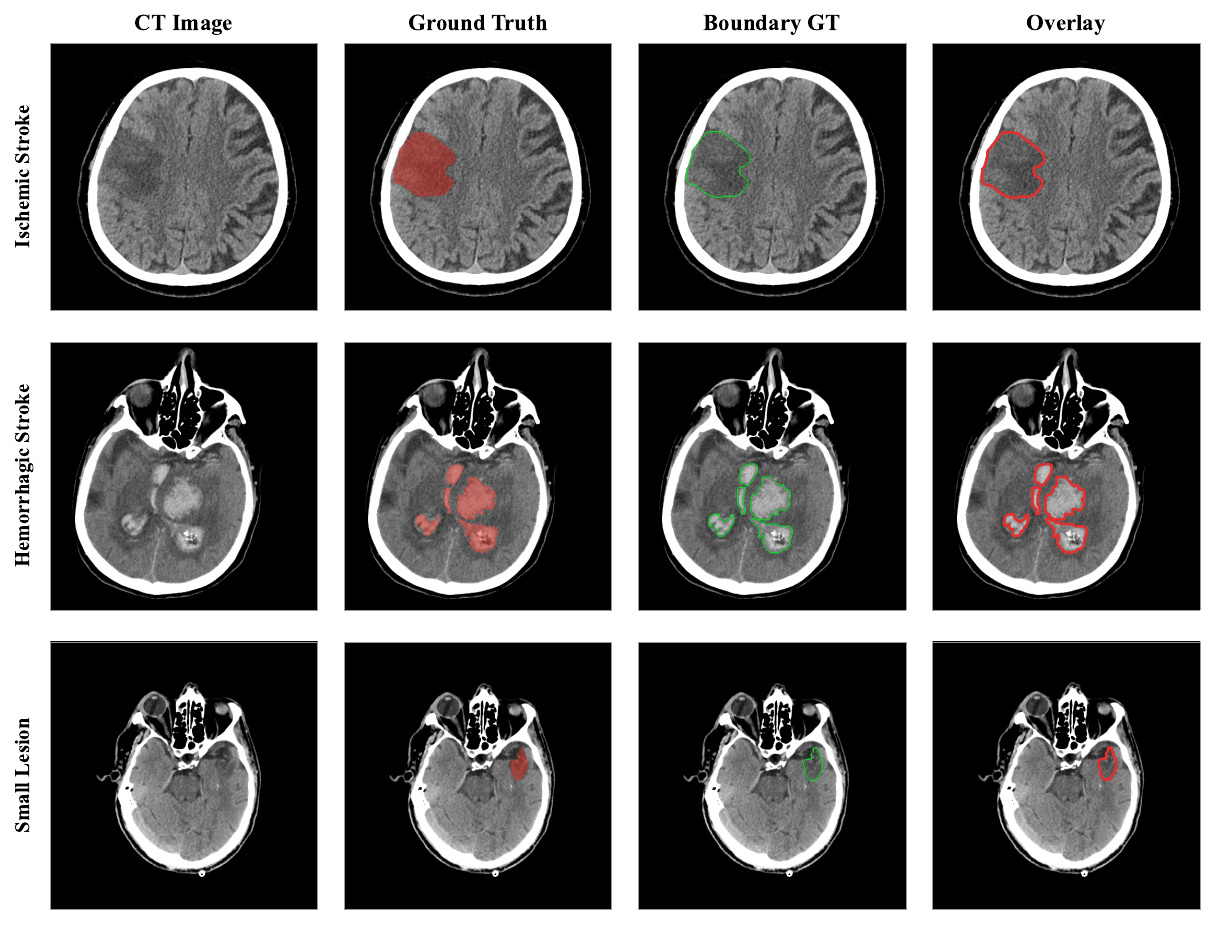}
    \caption{Qualitative visualization of stroke cases. Each row shows a different case type (ischemic, hemorrhagic, small lesion). Columns from left to right: CT image, ground truth mask overlay, boundary ground truth (morphological gradient), and contour overlay. The boundary ground truth highlights the challenging boundary regions that FSB-Net is designed to address.}
    \label{fig:qualitative}
\end{figure}

Fig.~\ref{fig:qualitative} presents representative segmentation results on cases from both stroke subtypes, including an ischemic case with subtle low-contrast boundaries and a hemorrhagic case with irregular margins. FSB-Net produces contours that more faithfully follow the ground-truth boundary, particularly for ischemic lesions where the boundary between lesion and normal tissue is subtle. The wavelet boundary features provide clear activation at lesion edges even when the spatial contrast is minimal, demonstrating the advantage of frequency-domain boundary modeling for low-contrast lesion segmentation.

\subsection{Computational Efficiency}
\label{sec:efficiency}

Table~\ref{tab:efficiency} compares computational requirements. FSB-Net achieves the best accuracy among all methods while maintaining a reasonable computational footprint with 27.4\,M parameters.

\begin{table}[!t]
\centering
\caption{Computational Efficiency Comparison}
\label{tab:efficiency}
\renewcommand{\arraystretch}{1.15}
\begin{tabular}{lccc}
\toprule
\textbf{Method} & \textbf{Params (M)} & \textbf{mDice (\%)} & \textbf{HD95 (px)} \\
\midrule
U-Net~\cite{ronneberger2015unet}           & 32.5  & 94.19 & 2.01 \\
UNet++~\cite{zhou2019unetpp}              & 26.1  & 93.75 & 2.70 \\
MANet~\cite{li2022manet}                  & 28.3  & 92.10 & 4.57 \\
DeepLabV3+~\cite{chen2018deeplabv3p}      & 26.7  & 93.78 & 2.11 \\
\midrule
\textbf{FSB-Net (Ours)}                   & 27.4  & \textbf{94.85} & \textbf{2.01} \\
\bottomrule
\end{tabular}
\end{table}

\section{Discussion}
\label{sec:discussion}

\textbf{Why frequency-domain boundary modeling works for stroke CT.}
The effectiveness of FSB-Net stems from the fundamental property that lesion boundaries produce high-frequency energy in the spectral domain. This is particularly relevant for stroke CT, where ischemic lesions appear as subtle density changes that are difficult to detect with spatial-domain edge operators. The DWT provides a natural multi-resolution decomposition that isolates boundary information across multiple scales and orientations (horizontal, vertical, diagonal), enabling the network to capture boundary features even when the spatial contrast is minimal.

\textbf{Bidirectional cross-attention enables mutual enhancement.}
The ablation on feature fusion strategies (Table~\ref{tab:ablation_attention}) confirms that bidirectional cross-attention outperforms simpler approaches. The boundary stream helps the spatial stream focus on edge regions, while the spatial stream provides contextual information that helps the boundary stream suppress false positives from noise and imaging artifacts common in CT scans.

\textbf{Handling heterogeneous stroke subtypes.}
The Brain Stroke CT Dataset contains both ischemic and hemorrhagic lesions with substantially different imaging characteristics. The multi-scale nature of the WBDH allows the network to adapt boundary detection to both subtle ischemic boundaries and the more conspicuous but irregular hemorrhagic margins. The consistent improvement across both subtypes suggests that frequency-domain boundary modeling is generalizable across different lesion appearances.

\textbf{Limitations.}
FSB-Net operates on 2D CT slices and does not exploit volumetric context from adjacent slices, which could improve segmentation consistency. The Haar wavelet, while computationally efficient, may not be optimal for curved boundary patterns; learned wavelet bases could improve performance. The current model does not distinguish between ischemic and hemorrhagic subtypes, which may be clinically relevant for treatment planning.

\section{Conclusion}
\label{sec:conclusion}

We presented FSB-Net, a frequency-spatial boundary network for brain stroke lesion segmentation in non-contrast CT. By integrating wavelet-based boundary detection, bidirectional frequency-spatial cross-attention, and a spectral boundary loss, FSB-Net achieves superior segmentation performance compared to established baselines. Experiments on the Brain Stroke CT Dataset demonstrate improvements in both overlap metrics (mDice, mIoU) and boundary precision (HD95), with the most pronounced gains in ischemic cases where low-contrast boundaries pose the greatest challenge. The frequency-domain approach to boundary modeling provides a principled alternative to spatial-domain edge operators, particularly suited to the low-contrast nature of stroke lesions on NCCT. Future work will extend FSB-Net to 3D volumetric segmentation and explore its application to other low-contrast lesion types.

\bibliographystyle{IEEEtran}
\bibliography{fsb_refs}

\end{document}